\pdfoutput=1

\documentclass[11pt]{article}

\usepackage{ACL2023}

\usepackage{times}
\usepackage{latexsym}

\usepackage[T1]{fontenc}

\usepackage[utf8]{inputenc}

\usepackage{microtype}

\usepackage{inconsolata}

\usepackage{todonotes}
\usepackage{graphicx}

\title{Interpreting and Mitigating Unwanted Uncertainty in LLMs
}

\author{
Tiasa Singha Roy\thanks{\ \ Equal contribution.},
\ Ayush Rajesh Jhaveri\footnotemark[1],
\ Ilias Triantafyllopoulos\footnotemark[1]\\
New York University \\
\texttt{\{ts5478, aj4332, it2190\}@nyu.edu} \\
}

\begin{document}
\maketitle
\begin{abstract}
Despite their impressive capabilities, Large Language Models (LLMs) exhibit unwanted uncertainty, a phenomenon where a model changes a previously correct answer into an incorrect one when re-prompted. This behavior undermines trust and poses serious risks in high-stakes domains. In this work, we investigate the mechanisms that drive this phenomenon. We adapt the Needle-In-A-Haystack retrieval framework and integrate a Flip-style re-evaluation prompt to simulate realistic answer-flipping scenarios. We find that retrieval heads are not primarily responsible for avoiding uncertainty. Instead, we identify a small set of non-retrieval attention heads that disproportionately attend to misleading tokens in uncertain contexts. Masking these heads yields significant improvements—reducing flip behavior by up to 15\% without introducing incoherence or overcorrection. However, when tested for downstream tasks, we observe tradeoffs with flip behavior. Our findings contribute to the growing field of mechanistic interpretability and present a simple yet effective technique for mitigating uncertainty-driven failure modes in LLMs.

\end{abstract}
\section{Introduction}

\begin{figure*}[t]
  \centering
  \includegraphics[width=0.9\linewidth]{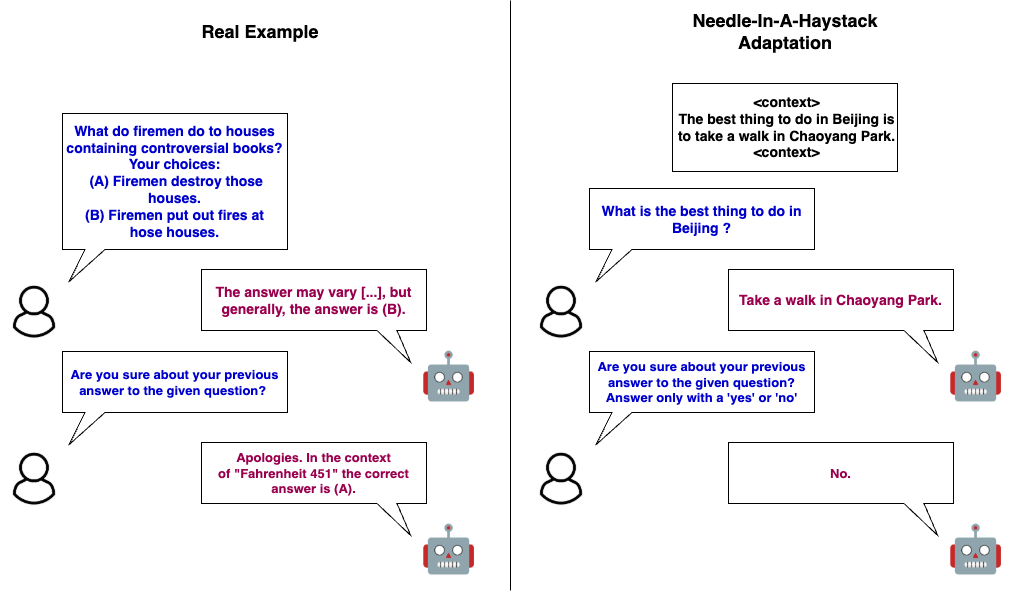}
  \caption{A visualization of the idea that is described in the paper. (Left) A real example that motivated our idea. The model answers a correct response on a multiple-choice question, but when it is asked if it is sure of its response, it changes its opinion to a false choice. (Right) An illustration of how we adapted this scenario to the Needle-In-A-Haystack task. We add the correct needle inside a large context, and then we ask the model to respond to a question that is answered by this needle. The model, again, responds correctly, but when it is asked to express its certainty, it changes its mind.}
  \label{motivation_diag}
\end{figure*}

Large Language Models (LLMs) such as GPT-4 and LLaMA-2 have achieved remarkable success across a wide range of NLP tasks. Yet despite their fluency and breadth of knowledge, they remain vulnerable to critical reliability issues, including hallucination \cite{Huang_2025}, overconfidence \cite{mielke-etal-2022-reducing}, and unwanted shifts in certainty under pressure \cite{sharma2023towards}. 
A particularly striking example of this fragility is shown in Figure 1 (left): when asked a factual multiple-choice question, the model initially selects the correct answer. 
However, when prompted to confirm its choice, it changes its mind—selecting an incorrect response it had previously rejected. 
This behavior—where models revise correct answers into hallucinated ones upon re-evaluation—is what we refer to as \textbf{unwanted uncertainty}.

The phenomenon of unwanted uncertainty has troubling implications. In high-stakes contexts like law or healthcare, models that second-guess correct answers or defer to user prompts may erode trust or even cause harm. 
Prior research has documented related behaviors such as hallucination \citep{farquhar2024detecting}, overconfident miscalibration \citep{leng2024taming}, and sycophantic response-shifting \citep{fanous2025syceval}, but these are often studied independently. 
Here, we bring them together by asking: which internal mechanisms cause LLMs to flip from correct to incorrect answers under questioning and how can we intervene?

To explore this, we design a controlled setup that adapts the Needle-in-a-Haystack retrieval benchmark \citep{kamradt2023needle} and integrates the FlipFlop re-evaluation paradigm \citep{laban2024surechallengingllmsleads}. 
As shown in Figure 1 (right), we embed a known correct answer (“needle”) into a long context and verify whether the model can retrieve and commit to this answer, even after being challenged. 
We then track the effect of masking specific attention heads to pinpoint those driving unwanted uncertainty.

Our findings reveal that this behavior is not mitigated by known retrieval heads \citep{wu2024retrieval}. 
Instead, we identify non-retrieval attention heads that consistently attend to misleading tokens when a model flips its answer from correct to incorrect. 
Targeted masking of these heads substantially improves consistency: reducing flip behavior by up to 15\% compared to no masking.

In sum, this work makes the following contributions:

\begin{itemize}
    \item We introduce a novel experimental setup to study unwanted uncertainty.
    \item We show that specific attention heads are responsible for answer-flipping behavior under uncertainty.
    \item We demonstrate that targeted head masking can mitigate this issue, improving model reliability without harming core performance.
\end{itemize}

\section{Related work}
Large language models often change or doubt a previously correct answer when challenged, a behavior frequently called sycophantic vacillation or the flip-flop effect. Recent studies \cite{laban2024surechallengingllmsleads} have quantified this phenomenon: in one multi-turn FlipFlop experiment across 9–10 models, LLMs flipped their answers 46\% of the time on average, with an accompanying accuracy drop of about 17\% from the first answer to the final answer. This effect has been observed in both cutting-edge proprietary models and open-source models \cite{xie2023ask}, indicating a universal tendency to vacillate under user skepticism. 

The survey by \citet{geng2023survey} suggests that confidence and correctness signals can be inferred from internal activation patterns in LLMs, with empirical findings indicating that certain layers and attention heads encode correctness-related features. In particular, \citet{beigi2024internalinspector} shows that contrastive learning over internal states—including attention and feed-forward activations—can effectively model confidence and distinguish correct from incorrect outputs. By leveraging such activation-level features, they show that confidence is not just a function of output logits but is embedded in layer-wise computation, potentially allowing for fine-grained uncertainty analysis.

This aligns with the finding from \citet{wu2024retrieval}, which identifies specialized attention heads—retrieval heads—that drive recall behavior in LLMs. \citet{olsson2022context} also shows that certain heads implement in-context induction algorithms. These findings illustrate that even within black-box LLMs like GPT-style transformers, individual heads can implement distinct, human-understandable functions. Such insights are foundational to understanding model behavior, as similar attention head circuits have been linked to models maintaining consistency across a dialogue \citep{zhao2024towards}. Based on these observations, we hypothesize that some attention heads play a similar role in modulating confidence or inducing uncertainty, especially in response to follow-ups. These heads may be causally implicated in the model's tendency to flip or second-guess an otherwise correct answer.

\section{Approach}




To isolate the uncertainty behavior, we extend the needle-in-a-haystack test by employing the flip experiment to evaluate the behavior of these heads when unwanted uncertainty occurs.

\begin{figure}[htbp]
    \centering
    \includegraphics[width=0.45\textwidth]{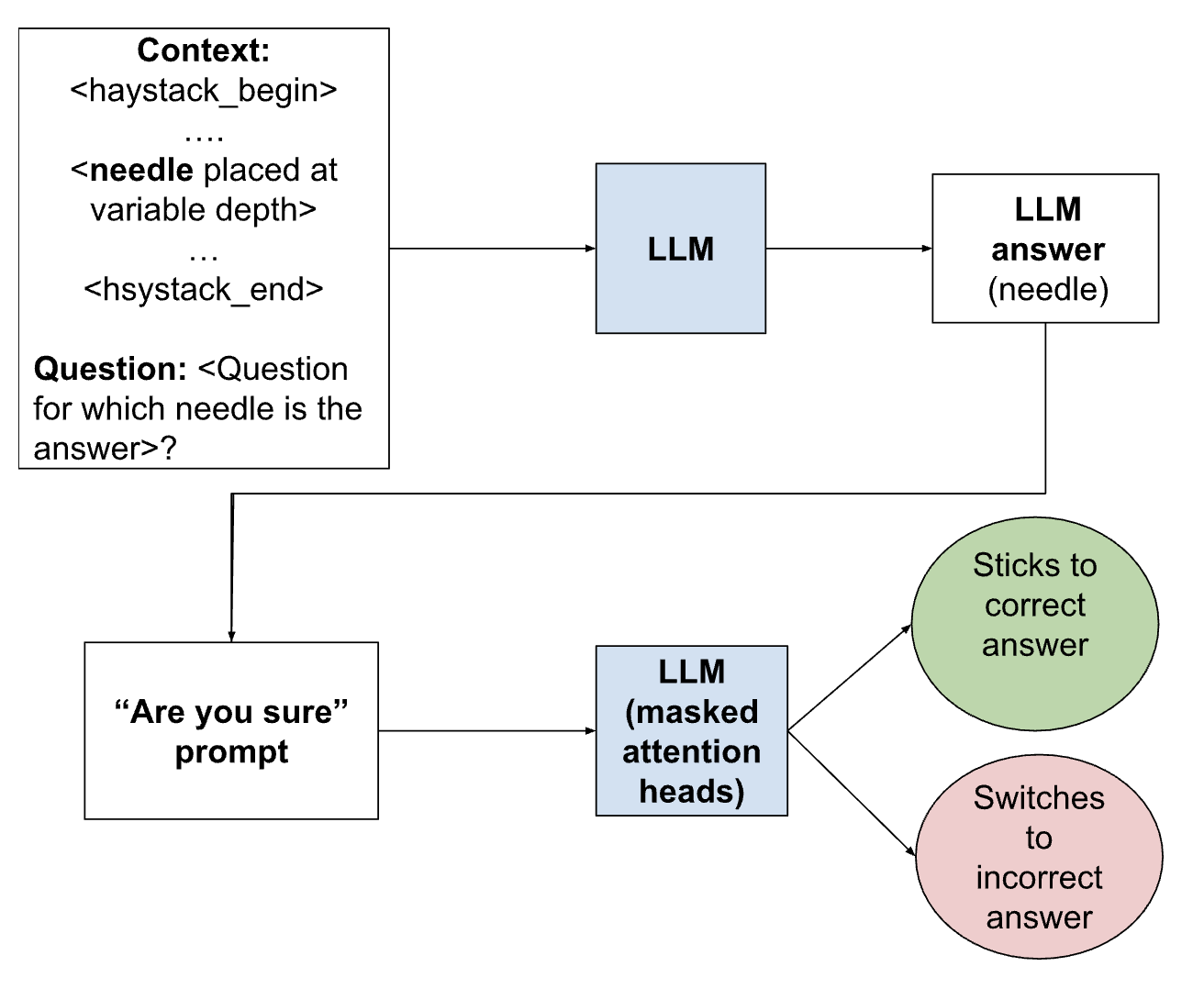}
    \caption{Flow diagram indicating the extension of the needle-in-a-haystack experiment to interpret unwanted uncertainty in LLMs.}
    \label{flow_diagram}
\end{figure}

We begin by prompting the LLM with a question and record its initial response. The model is then prompted to reevaluate its answer, where it either maintains or changes its response (see Fig.~\ref{flow_diagram}).

\subsection{Retrieval Heads Analysis}
To investigate retrieval head behavior, we analyze cases where the LLM switches to an incorrect answer. Specifically, we compare the effects of masking top retrieval heads against masking randomly selected heads for the re-evaluation prompt. If retrieval head masking does not lead to a higher rate of incorrect switches compared to random head masking, we can infer that retrieval heads are not primarily responsible for maintaining the model’s certainty in its responses. 

\subsection{Uncertainty Heads Exploration}
\label{section32}
We also analyze other attention heads to determine if they contribute to inducing uncertainty in LLMs. If specific attention heads show distinct activation patterns correlated with incorrect answer switches, we may attribute them as potential drivers of uncertainty in LLMs. 

To test this hypothesis, we re-purpose the retrieval score metric to compute head activations for 'yes'/'no' answer tokens. 

This is given by Activation Score, $h$, calculated as: 
\begin{equation}
    h_a = {\left|g_h \cap \{a\}\right|}
\end{equation}
where \( g_h \) is the set of tokens that head $h$ attends to most strongly at each decoding step, \( a \) is the answer token, which can be 'yes' or 'no'.

\begin{figure}[htbp]
    \centering
    \includegraphics[width=0.5\textwidth]{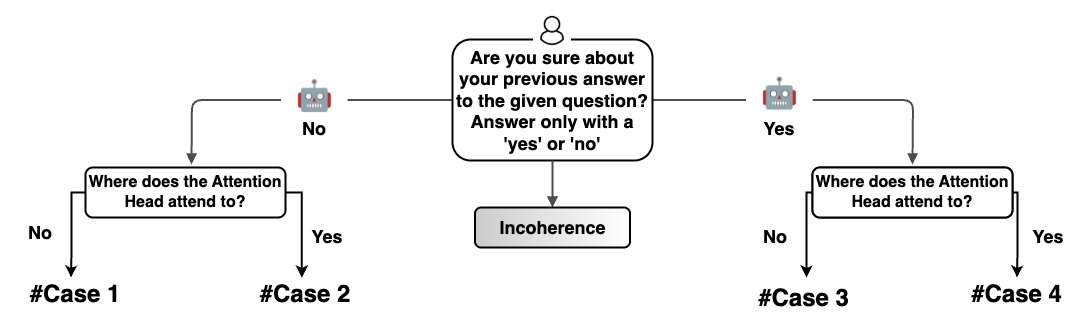}
    \caption{The response-head attention configurations. We consider our setup under the assumption that the response is always correct. The diagram for incorrect cases is symmetrical and explored further in our experiments.}
    \label{Experiment}
\end{figure}

To categorize attention behaviors, we define four response-head attention configurations (Cases 1–4), as visualized in Fig.~\ref{Experiment}. These are derived by conditioning on: (i) whether the model expressed confidence or uncertainty in the follow-up prompt, and (ii) whether each attention head attended to the answer token during decoding.

This case allows us to classify attention heads based on their activation patterns in different correctness-confidence contexts. We specifically check heads that consistently attend to the answer token when the model answers correctly, in order to identify cases of "unwanted" uncertainty,  where the model second-guesses a correct response.

To probe the functional role of these heads, we compare sets across conditions using three key operations. These operations combine the top heads of two different cases, $h_m$, $h_n$
\begin{itemize}
\item Intersection, which identifies heads common to both cases; this is given by $h_{m} \cap h_{n}$.

\item Difference, which isolates heads unique to one case (e.g., heads in Case 3 but not in Case 4); this is given by $h_{m} \setminus h_{n}$.

\item Union, which combines all heads active across both cases; this is given by $h_{m} \cup h_{n}$.
\end{itemize}

Among these, we hypothesize that the union of Case 1 and Case 2 captures attention heads involved in decision-making under uncertainty. These heads may encode conflicting signals that drive the model toward or away from expressing confidence, and may be central to the generation of flipped or incoherent responses.

To further investigate this, we isolate and analyze these heads under different experimental setups to better understand the attention mechanisms that contribute to unwanted uncertainty. To validate any observed correlations, we perform a causal test by masking these heads and examining whether the model continues to exhibit similar uncertainty-related behavior.

\section{Experimental Setup}

\subsection{Retrieval Heads Analysis}
\label{section41}

\textbf{Data} In \cite{kamradt2023needle}, a dataset is proposed consisting of question-answer pairs, where each answer (needle) is randomly embedded within unrelated long input texts (haystack) at various depths. We use a haystack length of up to 5,000 tokens to reduce GPU memory requirements. For retrieval head detection, we generate a total of 600 samples with 3 different needles. For the Flip extension test in Fig.~\ref{flow_diagram}, we use a different test needle with new contexts, yielding 400 samples. 

\textbf{Baseline Models} We select \textit{Llama-3.1-8B-Instruct} to perform our experiments as it is an open-source LLM that would enable experimentation with attention heads and has low computational requirements.

\textbf{Needle Insertion} We insert the needle into the haystack context without the \textit{begin\_of\_text} token, which was originally used in \cite{wu2024retrieval}. The rationale for this modification is discussed in Section~\ref{finding3}.

\textbf{Re-evaluation Prompt} We conduct our experiments using the following re-evaluation prompt: \textit{"Are you sure about your previous answer to the given question? Answer only with 'yes' or 'no'"}. This ensures that the LLM responds strictly with \textit{Yes} or \textit{No}. A \textit{Yes} answer token indicates that the LLM maintains its original (correct) answer, while a \textit{No} answer token signifies that it has switched to an incorrect answer.

\textbf{Head Masking} To evaluate LLM stability, we compare its responses when masking 10, 20, 30, and 50 top heads versus randomly selected heads. Here, \textbf{top heads} refer to attention heads with the highest retrieval scores. 
\footnote{Even if we consider "retrieval heads" the heads with higher than 0.5 score, we utilize the term "retrieval heads" in our experiments for denoting the heads with the highest score, regardless of passing the threshold of 0.5. Our masking strategy of top heads, masks all retrieval heads which we observe to be the top 2 heads for our baseline model. }

\textbf{Evaluation} We evaluate our approach using the following metrics:
\begin{itemize}
\itemsep0em 
    \item \textbf{Retrieval score:} frequency of an attention
head’s copy-paste behavior when answering questions that asks for raw information from the input. A retrieval score greater than or equal to 0.5 indicates a retrieval head. 
    \item \textbf{Recall score:} proportion of the needle that is part of the answer.
    \item \textbf{\% Yes responses:} a decrease in \textit{\% Yes} responses indicates lower stability in LLM responses.
\end{itemize}

\subsection{Uncertainty Heads Exploration}
\label{section42}

\textbf{Data} Building on the setup from Section~\ref{section41}, we expand the needle set to include 27 distinct needles—17 non-factual and 10 factual, while reducing variability in haystack length and needle depth. We consider a combination of non-factual needles to incorporate unseen synthetic, subjective content and factual needles which are rooted in real-world information. 
We make this change as we hypothesize that uncertainty heads should be more consistent across answer types, and less unhinged by context length and needle depth variations. This was needed in Section~\ref{section41} for detecting consistent retrieval capabilities.  Our setup yields 540 samples for detecting attention heads correlated with unwanted uncertainty. For mitigation testing, we use 5 (3 non-factual, 2 factual) unseen test needles in varied contexts, generating 400 samples.

\textbf{Baseline Models} As in Section~\ref{section41}, we use \textit{LLaMA-3.1-8B-Instruct} to ensure consistency across evaluations.

\textbf{Needle Insertion} As done for Section~\ref{section41}, we insert the needle into the haystack context without the \textit{begin\_of\_text} token, which was originally used in \cite{wu2024retrieval}. The rationale for this modification is discussed in Section~\ref{finding3}.

\textbf{Re-evaluation Prompt} We apply the same prompt as in Section~\ref{section41}: \textit{"Are you sure about your previous answer to the given question? Answer only with 'yes' or 'no'."}

\textbf{Head Masking} We test combinations of attention heads identified through the four response-head attention configurations described in Section~\ref{section32}. Specifically, we evaluate the effect of masking the intersection, difference and union of these cases on the \textit{Yes} and \textit{No} answer tokens. These configurations are compared against a no-mask baseline to isolate heads that causally contribute to unwanted uncertainty.

\textbf{Evaluation} We evaluate our approach using the following metric:
\begin{itemize}
\itemsep0em 
    \item \textbf{\% Yes responses:} A higher proportion indicates reduced uncertainty and reduced flip behavior. Decreases suggest heightened uncertainty and flip behavior under re-evaluation.
\end{itemize}

\section{Results and Discussion}

\subsection{Do Retrieval Heads contribute to LLM Stability?}

\begin{figure}[htbp]
    \centering
    \includegraphics[width=0.45\textwidth]{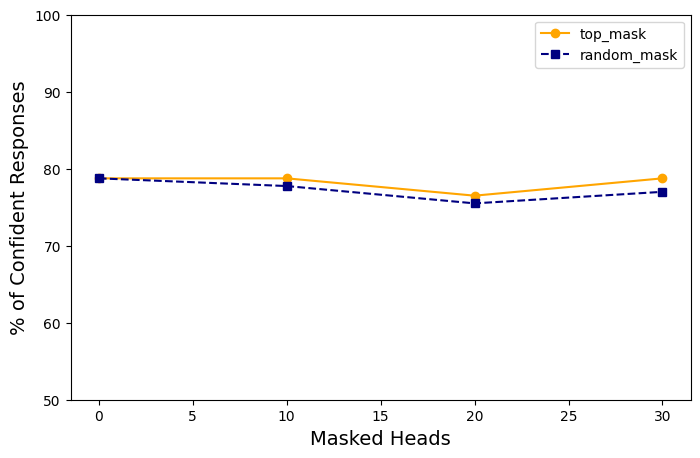}
    \caption{Effect on confident responses with masking for 400 samples. The unmasked case gives 315 \textit{Yes} responses and 85 \textit{No} responses, indicating an inherent model uncertainty.}
    \label{fig:2}
\end{figure}

Our initial experiments confirm that retrieval heads do not play a significant role in maintaining certainty in LLM responses. As shown in Figure~\ref{fig:2}, whether we mask up to 30 random heads or the top-ranked heads, we observe a similar gradual decline in Yes responses. Notably, the 30 top heads we mask mostly overlap with those having retrieval scores greater than 0.1, as shown in Table~\ref{heads_percentage}, where 34 heads exceed this threshold. This further supports our finding that retrieval heads do not substantially contribute to model certainty.

This implies the existence of a subset of uncertainty-associated heads, whose influence becomes more pronounced through different masking setups. To further validate this hypothesis, we seek to conduct a targeted analysis of individual head activations, focusing on identifying heads that correlate with increased uncertainty when left unmasked.



\subsection{Which Heads are responsible for the Unwanted Uncertainty?}

As explained in Section~\ref{section42}, we detect uncertainty heads and experiment with masking different logical combinations of them; combinations that, based on our observations, seem responsible for undesirable model behavior. The corresponding results are presented in Table \ref{main_results}.

One such combination involves masking the heads in Case 1 $\setminus$ Case 3. Both cases involve attention to the  token \textit{No} when the correct answer is \textit{Yes}. 
However, the difference is that in Case 1, the model outputs \textit{No}, while in Case 3, the model correctly outputs \textit{Yes}. This combination isolates the heads that specifically bias the model toward unwanted uncertainty. When we mask the top 10 heads from this set, we observe a modest but consistent \textit{Yes accuracy} increase of 1.3\%, from 67.5\% to 68.8\%. 

Our primary focus, though, is Case 1 $\cup$ Case 2, which captures all heads that pay attention to either \textit{Yes} or \textit{No} when the model incorrectly flips its response. This broader combination targets heads that focus on decision-making answer tokens in the case of unwanted uncertainty. 
Here, we observe that masking the top 5 heads in this set leads to a \textit{Yes accuracy} increase of 15\% (from 67.5\% to 82.5\%), while masking the top 10 still yields a substantial gain of 10.5\%. 
However, as we increase the number of masked heads beyond 10, the \textit{Yes accuracy} begins to drop sharply, with only 54.0\% \textit{Yes accuracy} at 20 masked heads. 
This suggests that excessive masking may remove helpful signal-processing mechanisms, showing that the number of these desired heads is actually small.

Further, we conducted targeted ablation experiments by masking a small number of specific heads that consistently scored highly across unwanted cases. 
These heads appeared to play a disproportionately large role in generating unwanted flip behavior. 
Masking only the head (11, 23) led to a modest gain of 1.5\%, indicating their influential role. 
When we added the head (17, 25), the \textit{Yes accuracy} increased to 72.3\%, a 4.8\% improvement. 
These results offer compelling evidence that a small number of attention heads disproportionately contribute to undesirable model behavior, and that selectively masking them can lead to meaningful performance gains with minimal disruption to overall functionality.

\begin{table}
\centering
    \begin{tabular}{l ccc}
        \textbf{Scenario} & \textbf{MH} & \textbf{Yes Accuracy} \\
        \hline
        No Mask & 0 & 67.5 \\
        \#Case 1 $\setminus$ \#Case 3 & 10 & 68.8 \\
        \#Case 1 $\cup$ \#Case 2 & 5 & \textbf{82.5} \\
        \#Case 1 $\cup$ \#Case 2 & 10 & 78.0 \\
        \#Case 1 $\cup$ \#Case 2 & 15 & 72.8 \\
        \#Case 1 $\cup$ \#Case 2 & 20 & 54.0 \\
        (11, 23) Head & 1 & 69.0 \\
        (11, 23), (17, 25) Heads & 2 & 72.3 \\
\hline
\end{tabular}
\caption{The main results of our experiments. MH stands for the number of Masked Heads for each row of results. Accuracy is measured as the percentage of "Yes" responses out of the total responses.}
\label{main_results}
\end{table}

\paragraph{Control Experiment:}

We have identified uncertainty heads that upon masking, increase the \% \textit{Yes} responses in the case that the correct answer was \textit{Yes}. Nevertheless, an alternate explanation for this behavior is that masking these heads might simply bias the model toward \textit{Yes} responses regardless of context.

To rule out this possibility, we design a scenario in which the model answers the question incorrectly, and hence, when re-prompted, the correct response should be \textit{No}. 
This was achieved by modifying the model conversation history with hand-crafted incorrect answers.

In this control setting, we evaluate the effect of masking the previously identified heads for the 5 test needles. 
Accuracy was now defined as the percentage of correct \textit{No} responses. 
Notably, we observed no degradation in performance. 
The accuracy remained stable at 100\%, even after masking. 
This result demonstrates that these heads do not simply promote \textit{No} responses. 
Instead, their influence appears to be highly context-dependent, specifically activated in situations where the model erroneously responds \textit{No} when it should answer \textit{Yes}.
We note that although the crafted needles contained cues that could plausibly mislead the model, the model consistently captured the invalidity of them. 
We further explore the implications of this in Section~\ref{section6}. 


\subsection{Additional Findings}

\subsubsection{Finding 1: There exist some non-retrieval heads responsible for coherent outputs}
\label{finding2}

\begin{table}
\centering
\begin{tabular}{ccc}
\hline
\textbf{Masked Heads}  & \textbf{IR - Ret} & \textbf{IR - Ran}\\
\hline
0 & 0 & 0\\
10 & 0 & 6\\
20 & 0 & 6\\
30 & 0 & 12\\
50 & 0 & 15\\
\hline
\end{tabular}
\caption{
The number of incoherent outputs with different attention head masking strategies when the model response expected is Yes or No. IR stands for the number of Incoherent Responses. Ret denotes the top retrieval heads, while Ran some random heads.}
\label{incoherent_counts}
\end{table}

During our experiments, we observe that in certain cases, the LLM produces incoherent outputs when responding to the re-evaluation prompt instead of a strict "Yes" or "No." As shown in Table~\ref{incoherent_counts}, this occurs when masking various numbers of random heads but not when masking top heads. This suggests that there is a subset of heads which, when unmasked, play a role in maintaining coherent outputs.

\subsubsection{Finding 2: Token identifiers in the middle of text give better retrieval scores}
\label{finding3}

Upon a detailed examination of the retrieval heads' codebase, we identified an undocumented implementation detail in the Needle-in-a-Haystack experiment: the needle was inserted into the context with a \textit{begin\_of\_text} token specific to each model. 
This detail, which was not mentioned in the original paper, may significantly impact the retrieval mechanism.
We hypothesize that the presence of this token acts as a "sign", guiding retrieval heads toward the correct information. 
To investigate this effect, we conduct the retrieval head detection experiment under two conditions: (a) the original setup, (b) a modified setup, where the \textit{begin\_of\_text} token is removed. As shown in Table~\ref{heads_percentage}, which presents the distribution of attention heads across different recall bands, retrieval scores drop significantly in the modified setup. This verifies the hypothesis that this token serves as a signal for retrieval heads. 

Further, Table~\ref{heads_ranking} reports the retrieval scores of the top 5 heads for both conditions. 
We observe that the ranking of the heads also changes, indicating that this token may also favor token-specific retrieval heads over response-specific retrieval heads. 
In other words, we assume that some heads originally detected as retrieval heads, actually perform \textit{begin\_of\_text} retrieval instead of answer retrieval, e.g. head (8-1). 

Therefore, we conduct all experiments without the \textit{begin\_of\_text} token for a fair analysis.

\begin{table}
\centering
\begin{tabular}{lcc}
\hline
\textbf{Score} & \textbf{\# Heads w. token} & \textbf{\# Heads wo token}\\
\hline
0 & 684 & 727\\
0-0.1 & 298 & 263\\
0.1-0.5 & 38 & 32 \\
0.5-1 & 4 & 2 \\
\hline
\end{tabular}
\caption{
The number of heads with retrieval scores within specific bands for the two cases: (a) 
w. token: When the initial \textit{begin\_of\_text} token is included at the beginning of the needle, (b) wo token: When this token is not included.}
\label{heads_percentage}
\end{table}

\begin{table}
\centering
\begin{tabular}{cc}
\hline
\textbf{Head - Score w. token} & \textbf{Head - Score wo token}\\
\hline
(15-30) - 0.899 & (15-30) - 0.895\\
(24-27) - 0.568 & (24-27) - 0.544\\
(8-1) - 0.531 & (16-1) - 0.486\\
(27-7) - 0.521 & (8-1) 0.452\\
(16-1) - 0.480 & (20-14) 0.441\\
\hline
\end{tabular}
\caption{
The top-5 retrieval heads for both original and modified - where the \textit{begin\_of\_text} token is removed - setups.}
\label{heads_ranking}
\end{table}

\subsubsection{Finding 3: Factual needles lead to higher \textit{Yes accuracy}}
\label{finding4}

\begin{figure}[htbp]
    \centering
    \includegraphics[width=0.48\textwidth]{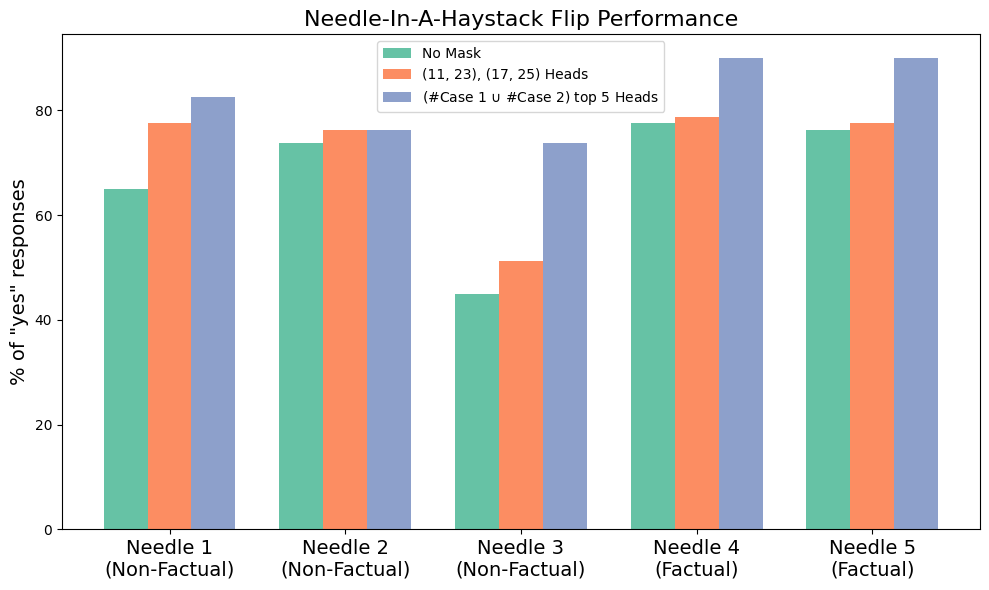}
    \caption{Accuracy (\% of "Yes" responses) for the different test needles.}
    \label{needle_scores}
\end{figure}

Our experiments reveal that factual statements exhibit significantly higher \textit{Yes accuracy} compared to subjective or unverifiable statements (Fig~\ref{needle_scores}). 
We tested recall performance using five different needle sentences: 3 non-factual and 2 factual. 
Our results show a clear trend—needles grounded in objective, factual information yield higher scores, while those containing opinions or unverifiable claims see a notable drop in performance.

This insight is crucial to our study on model uncertainty and flip behavior. 
If the model is more confident in retrieving verifiable information, uncertainty seems to be higher for unverifiable or subjective content.

\section{Influence of unwanted uncertainty heads on Downstream tasks}
\label{section6}

\subsection{Experiment Setup}

We previously identified candidate uncertainty heads and validated their role in mitigating uncertainty using the Needle-In-A-Haystack setup. To further test whether these heads truly contribute to uncertainty, we evaluate their impact across downstream tasks of varying difficulty.

We follow the same two-turn protocol as above: after the model answers a question, it is asked whether its response was correct. Ideally, the model should reply \textit{yes} when correct, and \textit{no} when incorrect. We compare performance across three settings: (i) no masking, (ii) masking heads (11, 23) and (17, 25), and (iii) masking Case 1 $\cup$ Case 2 heads.

\subsection{Experiment Results}

\begin{figure}[htbp]
    \centering
    \includegraphics[width=0.48\textwidth]{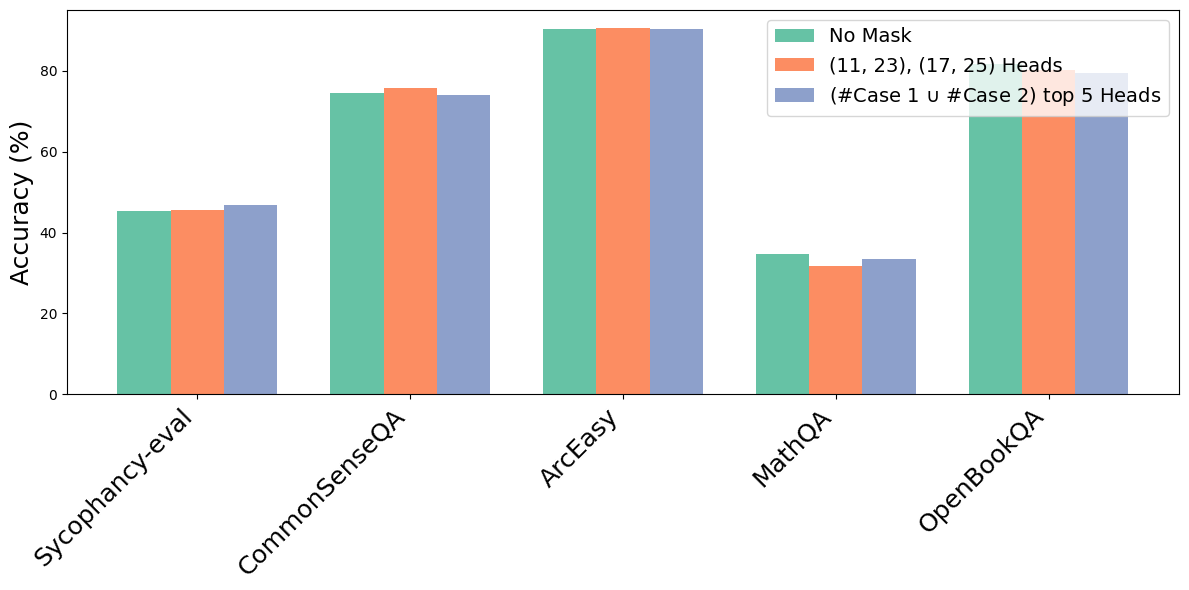}
    \caption{Downstream accuracy across different uncertainty head masking strategies.}
    \label{downstream_accuracy}
\end{figure}

Figure~\ref{downstream_accuracy} shows that first-turn accuracy is consistent across masking settings, aligning with our expectation that masking uncertainty heads would not impact the model's initial answer accuracy. Also, as expected, performance is higher on easier datasets (ArcEasy, OpenBookQA), moderate on CommonsenseQA and SycophancyEval, and lowest on the harder MathQA dataset.

\begin{figure}[htbp]
    \centering
    \includegraphics[width=0.48\textwidth]{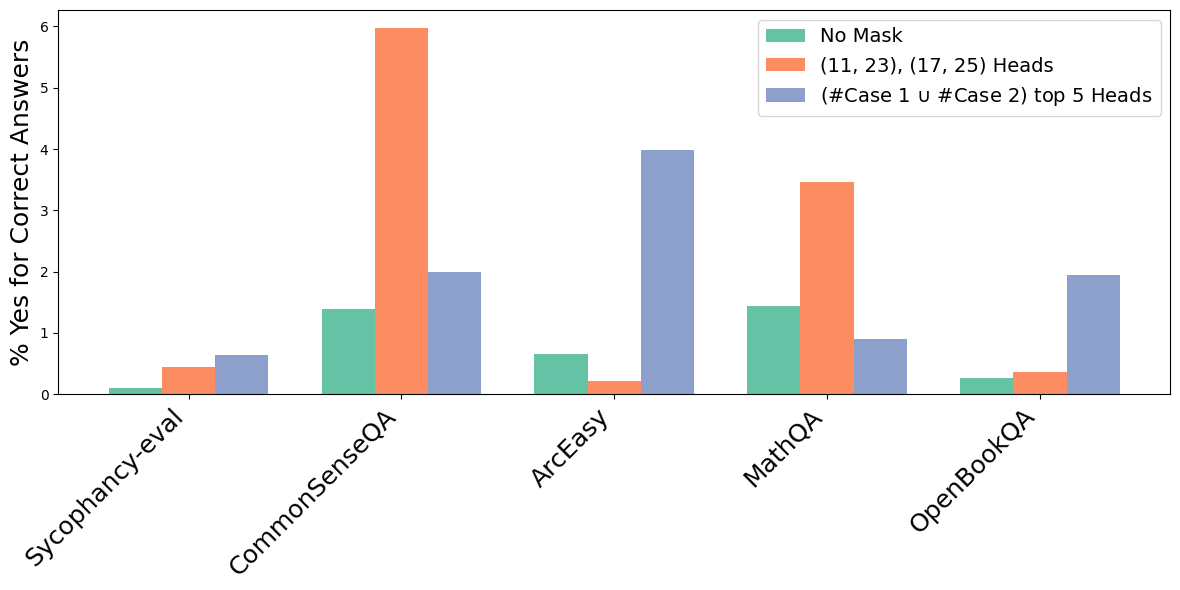}
    \caption{\% of "Yes" responses on second turn for \textbf{correct} answers across downstream tasks.}
    \label{downstream_correct}
\end{figure}

Figure~\ref{downstream_correct} reveals a clear trend for the Case 1 $\cup$ Case 2 masking: the rate of \textit{yes} responses for correct answers increases with dataset ease. This improvement is strongest for ArcEasy and OpenBookQA, modest for CommonsenseQA and SycophancyEval, but we observe a decrease for MathQA. This leads to the finding: \textit{masking uncertainty heads increases model confidence in low-uncertainty situations but may be counterproductive under high uncertainty}.

\begin{figure}[htbp]
    \centering
    \includegraphics[width=0.48\textwidth]{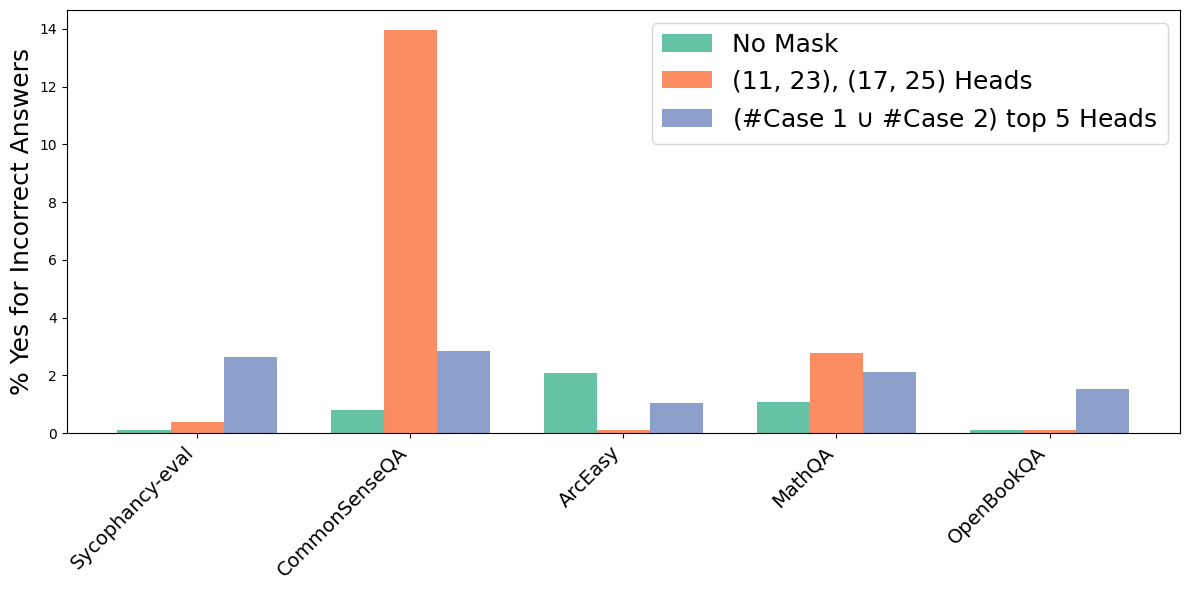}
    \caption{\% of "Yes" responses on second turn for \textbf{incorrect} answers across downstream tasks.}
    \label{downstream_incorrect}
\end{figure}

However, Figure~\ref{downstream_incorrect} highlights a potential drawback: the rate of \textit{yes} responses for incorrect answers also rises slightly across datasets. This raises a concern: \textit{does masking uncertainty heads increase overconfidence, reinforcing both correct and incorrect predictions?}

\section{Conclusion}

This work examines unwanted uncertainty in LLMs—cases where a model second-guesses a correct answer under re-evaluation. Using a setup that combines the Needle-in-a-Haystack framework with Flip-style prompts, we show that retrieval heads do not contribute to preventing flip behavior. Instead, a small set of non-retrieval attention heads consistently activate during incorrect flips.

By masking these heads, we reduce flip behavior by up to 15\% without harming baseline accuracy. Downstream evaluations reveal that masking uncertainty heads is effective only in cases of low uncertainty. These evaluations also expose a trade-off: improved confidence in high-certainty tasks, with slight overconfidence in low-certainty ones.

Our results show that unwanted uncertainty in LLMs is not a byproduct of retrieval failure, but rather a consequence of specific attention heads that influence the model’s decision-making under re-evaluation. By isolating and masking these heads, we can significantly reduce flip behavior while preserving overall accuracy. This suggests that certain internal mechanisms are not only interpretable but also controllable, offering a pathway toward more stable and trustworthy language models.

\section{Future Work}

Our findings highlight several directions for deepening the interpretability and robustness of uncertainty mechanisms in LLMs.

\textbf{Generalization}. Future work should test whether these uncertainty heads are present across different models (e.g., GPT-4, Claude, Mistral), architectures (Mixture-Of-Experts \citep{cai2024survey}), and tasks like multi-turn dialogue and safety-critical QA, instead of focusing entirely on the Needle-In-A-Haystack task.

\textbf{Incoherent Outputs}. Based on our findings in Section~\ref{finding2} that some non-retrieval heads yielded incoherent responses, future work could identify and characterize these heads.

\textbf{Better Uncertainty Head Detection}. Rather than focusing solely on token distributions, analyzing attention head output vectors may uncover latent shifts in internal states during flip behavior.


These directions move us toward more interpretable and controllable LLMs that can self-monitor and recover from uncertainty-induced errors.

\newpage

\bibliography{anthology,custom}
\bibliographystyle{acl_natbib}




\end{document}